\documentclass[10pt,journal,compsoc]{IEEEtran}
\ifCLASSOPTIONcompsoc
  \usepackage[nocompress]{cite}
\else
  \usepackage{cite}
\fi

\ifCLASSINFOpdf
\else
\fi

\usepackage{amssymb}
\usepackage{amsmath}
\usepackage{amsthm}

\usepackage[table]{xcolor}
\usepackage{bm}
\usepackage{graphicx}
\usepackage{booktabs}
\usepackage[bottom]{footmisc}
\usepackage{multirow}
\usepackage{threeparttable}
\usepackage{footmisc}
\usepackage{stfloats}
\usepackage{hyperref}

\newtheorem{prop}{Proposition}

\makeatletter
\newcommand\notsotiny{\@setfontsize\notsotiny{6.75}{7}}
\makeatother

\begin{document}

\title{A Note on Comparison of \textit{F}-measures}

\author{Wei Ju and Wenxin Jiang
\IEEEcompsocitemizethanks{\IEEEcompsocthanksitem W. Ju is with the Department
of Statistics, Northwestern University, Evanston,
IL, 60208.\protect \;
E-mail: wjz@u.northwestern.edu.
\IEEEcompsocthanksitem W. Jiang is with the Department
of Statistics, Northwestern University, Evanston,
IL, 60208.\protect \;
E-mail: wjiang@northwestern.edu.}}% <-this % stops an unwanted space


%A Preprint, December~2021. 
\markboth{\notsotiny \textrm{T\lowercase{his work has been submitted to the} IEEE \lowercase{for possible publication}. C\lowercase{opyright may be transferred without notice, after which this version may no longer be accessible}.}}%
{Shell \MakeLowercase{\textit{et al.}}: Bare Demo of IEEEtran.cls}

\IEEEtitleabstractindextext{%
\begin{abstract}
We comment on a recent TKDE paper \cite{c1} ``Linear Approximation of \textit{F}-measure for the Performance Evaluation of Classification Algorithms on Imbalanced Data Sets'',  and make  two improvements related to comparison of \textit{F}-measures for two prediction rules.
 
\end{abstract}

}

\maketitle

\IEEEdisplaynontitleabstractindextext

% \fi
\IEEEpeerreviewmaketitle

\IEEEraisesectionheading{\section{Introduction}\label{sec:introduction}}

\IEEEPARstart{F}{-measure} is a popular performance measure for classification algorithms, which compromises precision and recall. We found in a recent issue of TKDE Wong's paper \cite{c1} on statistical comparison of \textit{F}-measures for two algorithms, which is obviously an important problem. However, we found that there are two things in \cite{c1} that need improvement. 

\begin{itemize}

\item For each algorithm, Wong's variance formula in his Theorem 1 \cite{c1} has omitted the randomness of the weight for the recall. The correct implementation
should be via a delta method, which will lead to a different formula, see Takahashi,  Yamamoto,  Kuchiba and Koyama's Appendix C \cite{c2} for
a general formula allowing more than 2 classes. We conjecture that Takahashi \textit{et al}.'s formula \cite{c2} in the 2-class case should be equivalent to the ``JVESR formula'' (Janson and Vegelius 1981 \cite{c3};  Elston, Schroeder, and Rohjan  1982 \cite{c4}) that we use in this paper, which were introduced from different fields much earlier. These methods provide  analytic formulas  and do not need  \textit{k}-fold cross validation as in Wong \cite{c1}, for estimating the correlation between the recall and the precision. However, Takahashi \textit{et al}. \cite{c2} did not consider the comparison of two algorithms. 

\item Wong \cite{c1} does consider comparison of two algorithms, but has incorrectly assumed that they are independent in his Theorem 1, whereas they should really be correlated when applied to a common testing dataset (e.g., in Wong's Table 5 \cite{c1}).  In this paper, we use a formula from Ju \cite{c5} (Appendix A) for estimating the correlation between two different algorithms when operating on a same testing dataset. 

\end{itemize}

Our proposed method is therefore the combination of the use of  the ``JVESR formula'' for each algorithm, and the use of Ju's formula \cite{c5} (also see Proposition \ref{proposition}) for the between-algorithm correlation. The end result is that we can provide a correct way of comparing \textit{F}-measures for two algorithms, without the need of \textit{k}-fold cross validation as in \cite{c1}. The paper is organized as follows. We extend the ``JVESR formula'' in Section \ref{extension}. In Section \ref{experimental study}, we compare the numerical performance of the proposed method and Wong's method with the designed comparative experiments. Finally, we conclude and discuss possible future works in Section \ref{conclusion_future_work}.

%\subsection{Subsection Heading Here}
%Subsection text here.

\section{Extension of JVESR for Two \textit{F}-measures}
\label{extension}
In binary classification problems, the examples in a dataset (usually imbalanced) can be coded as $Z\in\{0,1\}$ and modeled as a random variable. The prediction results of examples in the dataset can be represented as $L_a\in\{0,1\}$,  respectively from the classification algorithm $a  \in\{1,2\}$ (in fact any finite set of   $a$ will be also okay).   
  
Assume $(Z,\{L_a\}),(Z_1,\{L_a\}_1),...,(Z_n,\{L_a\}_n),...$ are independent and identically distributed random vectors on $\{0,1\}\times\{0,1\}^2$. The sample average of all $f(Z_k,\{L_a\}_k)$ is denoted as 
\begin{equation}
E_nf(Z,\{L_a\}) \triangleq n^{-1} \sum_{k=1}^n f(Z_k, \{L_a\}_k).
\end{equation}
For $n=1,2,...,\infty,$  
  the performance of $L_a$ can be measured by the  \textit{F}-measure, which can be expressed as 
\begin{equation}
\tau_{na} \triangleq \frac{2 E_n(ZL_a)}{2 E_n(ZL_a)+E_n(L(1-Z_a))+E_n(Z(1-L_a))}.
\label{F_1_score}
\end{equation}
 
To compute the variance $\textrm{Var}(\tau_{n1}-\tau_{n2})$, we need to know $\textrm{Var}(\tau_{n1})$, $\textrm{Var}(\tau_{n2})$ and $\textrm{Cov}(\tau_{n1}, \tau_{n2})$. We extend the ``JVESR formula'' for computing the covariance of $F$-measures for two algorithms.

\begin{prop} (Covariance formula) \label{proposition} Denote $\kappa_{na}   \triangleq \tau_{na}^{-1} -1 $. Then for any $a,b\in\{1,2\}$, the large $n$ asymptotic covariance  
  \begin{equation}
   \begin{split}
\textrm{Cov}\big(\tau_{na} , \tau_{nb} \big) &= \frac{ \tau_{ a}^2\tau_{b}^2 }{n2^2 E(ZL_a)E(ZL_b)} \times \\
&\;\;\big[ \; \textrm{Cov}\{L_a(1-Z)+Z(1-L_a), L_b(1-Z)\} \\
&+  \textrm{Cov}\{L_a(1-Z)+Z(1-L_a), Z(1-L_b)\} \\
&-  \kappa_{a}  \cdot \textrm{Cov}\{ZL_a, L_b(1-Z)+Z(1-L_b)\}   \\
&-  \kappa_{b}  \cdot \textrm{Cov}\{ ZL_b, L_a(1-Z)+Z(1-L_a)\}  \\
&+  \kappa_{a}  \kappa_{b}  \cdot\textrm{Cov}\{ ZL_a, ZL_b\} \;\big],
\end{split}
\label{cov}
\end{equation}

\noindent where $\tau_{ a} = \lim_{n \to \infty} \tau_{na}$, $\tau_{b} = \lim_{n \to \infty} \tau_{nb}$, $\kappa_{a} = \lim_{n \to \infty} \\ \kappa_{na}$ and $\kappa_{b} = \lim_{n \to \infty} \kappa_{nb}$.
\end{prop}   

When $a=b$, the covariance in \eqref{cov} becomes the variance, which can be shown by tedious algebra to coincide with the ``JVESR formula'' (Janson and Vegelius 1981 \cite{c3};  Elston, Schroeder, and Rohjan  1982 \cite{c4}).  So we in fact are presenting an extension of the ``JVESR formula'' for multiple algorithms.

To apply the formulas in this Proposition, we can estimate the $\tau_a, \kappa_a, E$ and $\textrm{Cov}$ by the sample analogues. To compare  sample $F$-measure  $\Bar{f}_{a} = \tau_{na}$ for two algorithms $a=1,2$, this Proposition allows us to compute the following quantities in Table \textcolor{black}{\ref{comparison_results}} later, according to our proposed extension of the JVESR method:
\begin{equation}
    \textrm{Var}(\bar f_a)=\textrm{Cov}(\bar f_a,\bar f_a), \;\;\; \textrm{for}\; a=1,2,
\end{equation}
\begin{equation}
     \textrm{Corr}(\Bar{f}_1 , \Bar{f}_2) =\frac{\textrm{Cov}(\Bar{f}_1 , \Bar{f}_2) }{ \sqrt{ \textrm{Var}(\Bar{f}_1) }\sqrt{ \textrm{Var}( \Bar{f}_2)}},
\end{equation}
\begin{equation}
\begin{split}
    \textrm{Var}(\Bar{f}_1 - \Bar{f}_2) &= \textrm{Cov}(\Bar{f}_1,\Bar{f}_1) + \textrm{Cov}(\Bar{f}_2,\Bar{f}_2) \\
    & - 2 \textrm{Cov}(\Bar{f}_1, \Bar{f}_2), 
\label{var_f1-2}
\end{split}
\end{equation}
\begin{equation}
z=\frac{\Bar{f}_1 - \Bar{f}_2 }{ \sqrt{ \textrm{Var}(\Bar{f}_1 - \Bar{f}_2) }}. 
\end{equation}

\begin{figure*}[bp]
    \centering
        \includegraphics[scale=0.78]{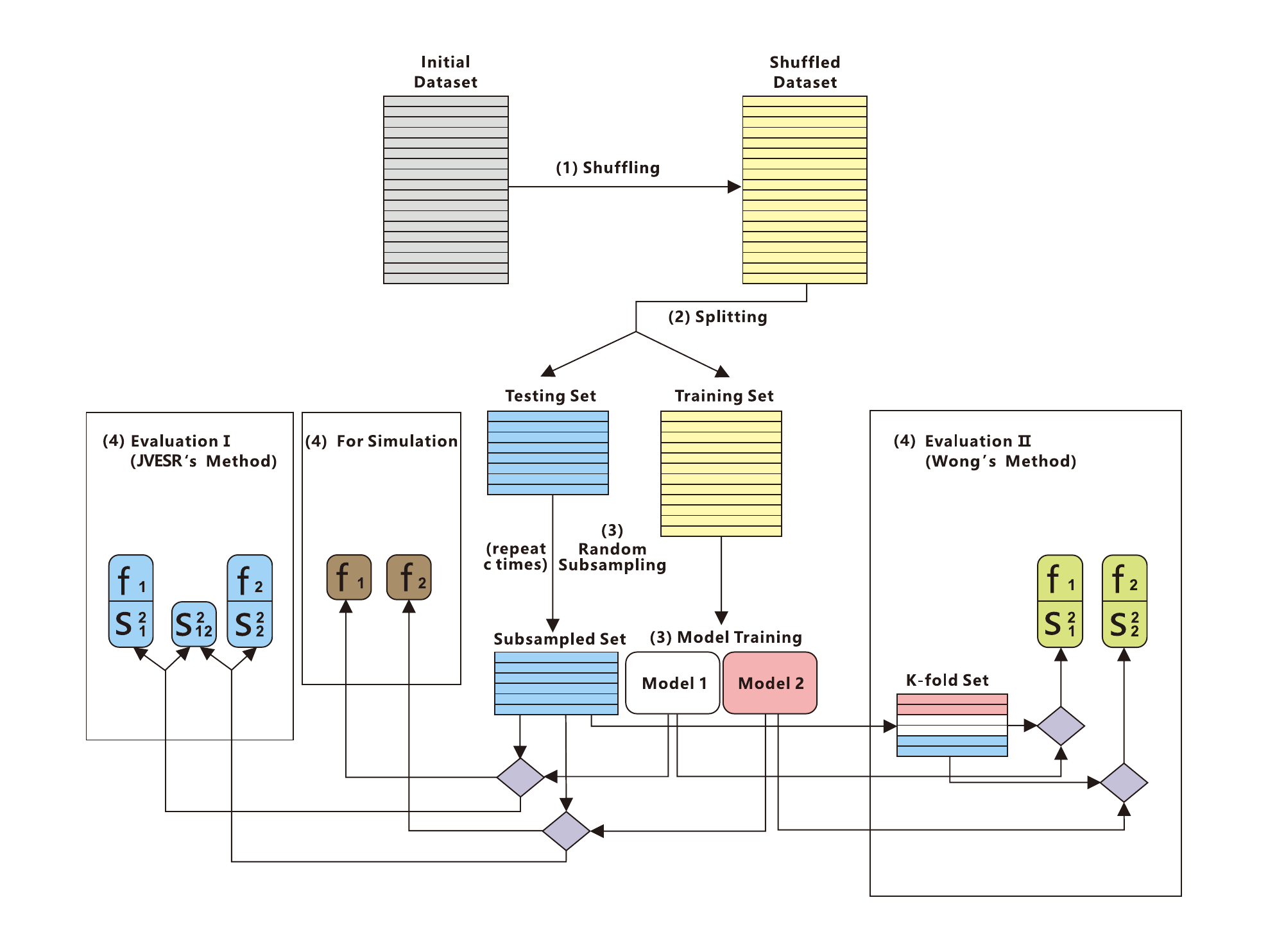}
    \caption{Illustration of the Comparative Experiments. We first shuffle the initial dataset, and then split the shuffled dataset into training set and testing set, followed by model training and random subsampling the testing set for evaluation and simulation procedures. We repeat the subsampling, evaluation and simulation procedures $c$ times. In the Figure, $s_i^2$ represents the estimated sample variance $\textrm{Var}(\Bar{f}_i)$, and $s_{12}^2$ represents the sample covariance $\textrm{Cov}(\Bar{f}_1, \Bar{f}_2)$.}
    \label{fig:Comparison}
\end{figure*}

%\bigbreak
\noindent \textbf{Proof of Proposition 1.} \\
For $i\in\{1,2\}$, let
\begin{equation}
\kappa_{ni}   \triangleq \tau_{ni}^{-1} -1 = \frac{E_n(L_i(1-Z)+Z(1-L_i))}{2 E_n(ZL_i)}.
\label{kappa}
\end{equation}

In order to apply the delta method, we need to differentiate $\kappa_{ni} $, 
\begin{equation}
\begin{split}
\mathrm{d}\kappa_{ni}   & = \frac{\mathrm{d}E_n(L_i(1-Z)+Z(1-L_i))}{2 E_n(ZL_i)}\\
& \;\;\;\; - \frac{E_n(L_i(1-Z)+Z(1-L_i))\cdot \mathrm{d}E_n(ZL_i)}{2 E_n^2(ZL_i)} \\
& \approx \frac{\mathrm{d}E_n(L_i(1-Z)+Z(1-L_i))-\kappa_{ni}  \cdot\mathrm{d}E_n(ZL_i)}{2 E(ZL_i)},
\end{split} 
\label{differentiate}
\end{equation}
where $E=E_{\infty}$.

With \eqref{differentiate} and the delta method, for any $a,b\in\{1,2\}$, we have,
\begin{equation}
\begin{split}
\textrm{Cov}\big(\kappa_{na}  , \kappa_{nb} \big) &= \frac{ 1 }{n2^2 E(ZL_a)E(ZL_b)}\times \\
& \;\;[\; \textrm{Cov}\{L_a(1-Z)+Z(1-L_a), L_b(1-Z)\} \\
&+  \textrm{Cov}\{L_a(1-Z)+Z(1-L_a), Z(1-L_b)\} \\
&-  \kappa_{a}  \cdot \textrm{Cov}\{ZL_a, L_b(1-Z)+Z(1-L_b)\}   \\
&-  \kappa_{b}  \cdot \textrm{Cov}\{ ZL_b, L_a(1-Z)+Z(1-L_a)\}  \\
&+  \kappa_{a}  \kappa_{b}  \cdot\textrm{Cov}\{ ZL_a, ZL_b\} \;].
\end{split}
\label{cov_kappa}
\end{equation}

The object $\kappa_{ni}  $ in \eqref{kappa} can also be differentiated as $\mathrm{d}\kappa_{ni}  \approx -\tau_{i}^{-2} \mathrm{d}\tau_{ni} $, and thus we have $\mathrm{d}\tau_{ni}  \approx -\tau_{ i}^2  \mathrm{d}\kappa_{ni} $. By applying the delta method again, the covariance of $\tau_{na}$ and $\tau_{nb}$ can be calculated as 
\begin{equation}
\begin{split}
   \textrm{Cov}\big(\tau_{na} , \tau_{nb} \big) =  \tau_{ a}^2  \tau_{ b}^2\cdot\textrm{Cov}\big(\kappa_{na} , \kappa_{nb} \big) ,
\end{split}
\label{cov_tau}
\end{equation}
which leads to the proof.
Q.E.D.

\begin{table*}[bp]
  \centering
  \scalebox{1.08}{
  \begin{threeparttable}[b]
  \caption{Results of Comparing JVESR Method With Wong's Method on the Three Datasets.}
  \label{comparison_results}
  
  \begin{tabular}{l|l|c|c|c|c|c|c}\toprule \midrule
  \textbf{Dataset} & \textbf{Metric} & \multicolumn{2}{c|}{\textbf{Simulation}} & \multicolumn{2}{c|}{\textbf{JVESR Method \cite{c3},\cite{c4}}} & \multicolumn{2}{c}{\textbf{Wong’s Method} \cite{c1}} \\ \midrule 
  \multicolumn{2}{c|}{(Algorithm)} & 1-NN & RF & 1-NN & RF & 1-NN & RF  \\ \midrule 
  \multirow{5}{*}{Abalone} & $\Bar{f}_{i^{\dagger}}$ & 0.209 & 0.138 & 0.209 & 0.138 & 0.209 & 0.138 \\ 
  & \textrm{Var}($\Bar{f}_i$) & 0.00241 & 0.00278 & 0.00237 & 0.00267& 0.00230 & 0.00235 \\ \cmidrule{3-8}
  & \textrm{Corr}($\Bar{f}_1, \Bar{f}_2$) & \multicolumn{2}{c|}{0.347} & \multicolumn{2}{c|}{0.348} & \multicolumn{2}{c}{0}  \\
  & \textrm{Var}($\Bar{f}_1 - \Bar{f}_2$) & \multicolumn{2}{c|}{0.00339} & \multicolumn{2}{c|}{0.00326} & \multicolumn{2}{c}{0.00465}  \\
  & $z$-statistic & \multicolumn{2}{c|}{1.22} & \multicolumn{2}{c|}{1.27} & \multicolumn{2}{c}{1.09}  \\
  \midrule 
  \multirow{5}{*}{White-wine} & $\Bar{f}_i$ & 0.347 & 0.445 & 0.347 & 0.445 & 0.347 & 0.445 \\ 
  & \textrm{Var}($\Bar{f}_i$) & 0.00276 & 0.00373 & 0.00287 & 0.00393& 0.00273 & 0.00264 \\ \cmidrule{3-8}
  & \textrm{Corr}($\Bar{f}_1, \Bar{f}_2$) & \multicolumn{2}{c|}{0.744} & \multicolumn{2}{c|}{0.746} & \multicolumn{2}{c}{0}  \\
  & \textrm{Var}($\Bar{f}_1 - \Bar{f}_2$) & \multicolumn{2}{c|}{0.00171} & \multicolumn{2}{c|}{0.00180} & \multicolumn{2}{c}{0.00537}  \\
  & $z$-statistic & \multicolumn{2}{c|}{-2.38} & \multicolumn{2}{c|}{-2.42} & \multicolumn{2}{c}{-1.38}  \\
  \midrule 
  \multirow{5}{*}{Seismic} & $\Bar{f}_i$ & 0.145 & 0.052 & 0.145 & 0.052 & 0.145 & 0.052 \\ 
  & \textrm{Var}($\Bar{f}_i$) & 0.00164 & 0.00070 & 0.00176 & 0.00127& 0.00173 & 0.00120 \\ \cmidrule{3-8}
  & \textrm{Corr}($\Bar{f}_1, \Bar{f}_2$) & \multicolumn{2}{c|}{0.342} & \multicolumn{2}{c|}{0.412} & \multicolumn{2}{c}{0}  \\
  & \textrm{Var}($\Bar{f}_1 - \Bar{f}_2$) & \multicolumn{2}{c|}{0.00161} & \multicolumn{2}{c|}{0.00179} & \multicolumn{2}{c}{0.00293}  \\
  & $z$-statistic & \multicolumn{2}{c|}{2.33} & \multicolumn{2}{c|}{2.16} & \multicolumn{2}{c}{1.75}  \\
  \midrule \bottomrule
  \end{tabular}
  \begin{tablenotes}\footnotesize
  \item[$\dagger$] $i=1$ if the algorithm is 1-NN. Otherwise, $i=2$.
  \end{tablenotes}
  \end{threeparttable}}
\end{table*} 

\section {Experimental Study}
\label{experimental study}
To compare the performance of the extended JVESR method and Wong's method, we propose a framework of comparative experiments, and a high-level overview of this framework is shown in Figure \ref{fig:Comparison}.
We choose 3 datasets (Abalone \cite{c6}, White-wine \cite{c7} and Seismic \cite{c8}) from the UCI data repository \cite{c9} for evaluating the model performance. For each of the three original datasets, we save a subset as training data, and apply  1-NN and RF (Random Forest), to obtain two prediction rules $L_1\in\{0,1\}$ and $L_2\in\{0,1\}$, respectively, for a  label $Z\in\{0,1\} $ defined according to Wong \cite{c1}.  The remaining data form a ``population"  from which the testing datasets, each of size $n=1000$, are  subsampled with replacement for $c=1200$ times.
On each subsampled testing dataset, the $F$-measures $\bar f_1$ and $\bar f_2$ of prediction rules $L_1$ and $L_2$, respectively,
are computed and compared. Our code is publicly available.\footnote{\label{rcode}\href{https://github.com/teniscape/comparison-of-methods-for-estimating-F-measures}{https://github.com/teniscape/comparison-of-methods-for-estimating-F-measures.}}

In Table \textcolor{black}{\ref{comparison_results}}, both JVESR and Wong's methods are applied for variance computation, based on testing datasets, each of size $n=1000$. Entries in Table \textcolor{black}{\ref{comparison_results}} represent the average values over multiple testing datasets. For both methods, a small percentage of these $c=1200$ testing datasets lead to infinite variance, and are therefore excluded from the average. The exact number of testing datasets we use is denoted as $c^{\scalebox{0.95}[0.7]{-}}$, and the values of $c^{\scalebox{0.95}[0.7]{-}}$ for datasets Abalone, White-wine and Seismic are $1137, 1173$ and $909$, respectively, for which the 
computed variances are finite for both JVESR and Wong’s methods.
(In this regard, more testing datasets are excluded for Wong’s method,
especially for the Abalone and White-wine datasets.) The ``simulated'' variance, correlation, etc., are computed with the $c^{\scalebox{0.95}[0.7]{-}}$ pairs of $(\bar f_1, \bar f_2)$ obtained from the $c^{\scalebox{0.95}[0.7]{-}}$ testing datasets. They approximately represent the ``true values'' for the  variance, correlation, etc. 

For the \textrm{Corr}($\Bar{f}_1, \Bar{f}_2$) using the JVESR method, we report the averaged value of all the $c^{\scalebox{0.95}[0.7]{-}}$ correlations between $\Bar{f}_1$ and $\Bar{f}_2$ of the algorithms tested on the subsamples, each being estimated from the covariance formula proposed in Proposition \ref{proposition}. The simulated \textrm{Corr}($\Bar{f}_1, \Bar{f}_2$) is computed as the exact correlation between the $c^{\scalebox{0.95}[0.7]{-}}$ $\Bar{f}_1$'s and $\Bar{f}_2$'s of the algorithms tested on the subsamples. In Table~\textcolor{black}{\ref{comparison_results}}, the \textrm{Corr}($\Bar{f}_1, \Bar{f}_2$) computed by the JVESR method is pretty close to the simulated correlation between $\Bar{f}_1$ and $\Bar{f}_2$ for all the three datasets, and this further verifies that the dependency of algorithms shouldn't be ignored when comparing the algorithms tested on the same dataset.

As shown in Table~\textcolor{black}{\ref{comparison_results}}, we present the $z$-statistic values for pairwise comparison of different methods, averaged over $c^{\scalebox{0.95}[0.7]{-}}$ testing datasets. It is possible for us to make quite opposite decision for pairwise comparison if the estimation of \textrm{Var}($\Bar{f}_1 - \Bar{f}_2$) is inaccurate. For example, for the White-wine and Seismic sets, the simulated $z$-statistic and and the $z$-statistic computed by the JVESR method both indicate that typically we have to reject the null hypothesis at the significance level 0.05 for comparing the two algorithms on both datasets. However, with the same significance level, the $z$-statistics computed with Wong's method indicate that we can typically accept the null hypothesis.

The ways JVESR and Wong used to compute the \textit{F}-measure are essentially the same, although Wong first calculates the recall and precision for data in each fold and then average them to calculate the final \textit{F}-measure for each algorithm. 
As a result, the \textit{F}-measures computed for the same algorithm on the same dataset shown in Table~\textcolor{black}{\ref{comparison_results}} are the same. Although for most of the cases, the variances of \textit{F}-measure computed using Wong's method are not accurate, we do find that they are close to the simulated variances in some cases, e.g., the \textrm{Var}($\Bar{f}_1$) for White-wine data. Although we have pointed out that the method Wong used to compute the variance of $\Bar{f}_i$ is not totally right, sometimes the computed variance can still be pretty close to the simulated result.

When testing the RF algorithm on the Seismic dataset, we find that the number of true positive (\textit{TP}) examples for most of the subsampled sets is less than five, and thus the large-sample condition is not satisfied for those cases. The precise asymptotics in the estimation of the variance in JVESR method is probably impaired in this situation, and this makes the estimated \textrm{Var}($\Bar{f}_2$) not so accurate as the estimations for the other two datasets. The estimated variance obtained with Wong's method in this case is also larger than the simulated variance. Even though the estimated varianc-\\ es for RF algorithm on Seismic data are not so accurate for either method, we still get more accurate $z$-statistics for pairwise comparison when applying the JVESR method since the correlation between the two algorithms can be naturally incorporated by our variance formula \eqref{var_f1-2}.

We notice that the $\Bar{f}_2$ for Seismic data is close to boundary 0, and this may be another possible reason ($\textit{F}$-measure being close to values 0 or 1) why it doesn't lead to a good asymptotic result when using JVESR method to estimate the \textrm{Var}($\Bar{f}_2$) on Seismic data. We leave for future work the more thorough understanding of this interesting phenomenon that possibly caused by small number of \textit{TP} examples and extreme $\textit{F}$-measure, and the finding of possible methods to fix it.

\section{Conclusions}
\label{conclusion_future_work}
In this paper, we improve Wong's method \cite{c1} for comparing two \textit{F}-measures, by allowing correlated prediction rules and applying the variance formulas of JVESR \cite{c3,c4}. Our proposed framework of comparative experiments can be extended to compare multiple algorithms on multiple datasets, which is an ongoing work.

Finally, we comment that by comparison of two algorithms, we really mean comparison of the testing data performances of the two rules learned from the same training data, as generated by the two different classification methods. In standard error computations of both JVESR and Wong, variance of testing data is incorporated but not the variation of the classification rules learning from possibly different training data. The latter is more difficult, since its variance does not relate to standard estimation of a low dimensional parameter from training data, such as in Logistic Regression, but is involving many parameters possibly incompletely optimized (such as in Random Forest) or a completely nonparametric method (such as in Nearest Neighbor).

%\ifCLASSOPTIONcompsoc
 % \section*{Acknowledgment}
%\else
 % \section*{Acknowledgment}
%\fi

%The authors would like to thank the UCI Machine Learning Repository for hosting the datasets.

\ifCLASSOPTIONcaptionsoff
  \newpage
\fi



%\bibliographystyle{IEEEtran}

\begin{thebibliography}{1}

\bibitem{c1}
T.~Wong, Linear approximation of \textit{F}-measure for the performance evaluation of classification algorithms on imbalanced data sets, \emph{IEEE Transactions on Knowledge and Data Engineering}, april, 2020.

\bibitem{c2}
K. Takahashi, K. Yamamoto, A. Kuchiba, and T. Koyama, Confidence interval for micro-averaged \textit{F}$_1$ and macro-averaged \textit{F}$_1$ scores, \emph{Applied Intelligence}, july, 2021.

\bibitem{c3}
S. Janson and J. Vegelius, Measures of ecological association, \emph{Oecologia}, vol. 49, pp. 371–376, 1981.

\bibitem{c4}
R. C. Elston, S. R. Schroeder, and J. Rohjan, Measures of observer agreement when binomial data are collected in free operant situations, \emph{Journal of Behavioral Assessment}, vol. 4, pp. 299–310, 1982.

\bibitem{c5}
W. Ju, Ensembling and data selection for neural language models, and analysis of \textit{F}-measure, Ph.D. dissertation, Northwestern University, 2021.

\bibitem{c6}
W. J. Nash, T. L. Sellers, S. R. Talbot, A. J. Cawthorn, and W. B. Ford, The population biology of abalone (\_haliotis\_species) in tasmania. i. blacklip abalone (\_h. rubra\_) from the north coast and islands of bass strait, \emph{Sea Fisheries Division, Technical Report}, no. 48, 1994.

\bibitem{c7}
P. Cortez, A. Cerdeira, F. Almeida, T. Matos, and J. Reis, Modeling wine preferences by data mining from physicochemical properties, \emph{Decision Support Systems, Elsevier}, vol. 47, no. 4, pp. 547–553, 2009.

\bibitem{c8}
M. Sikora and Ł. Wróbel, Application of rule induction algorithms for analysis of data collected by seismic hazard monitoring systems in coal mines, \emph{Archives of Mining Sciences}, vol. 55, no. 1, pp. 91–114, 2010.

\bibitem{c9}
D. Dua and C. Graff, UCI machine learning repository, Irvine, CA: University of California, School of Information and Computer Sciences, 2019.


\end{thebibliography}

\end{document}